\documentclass[a4paper, 12pt]{article}

\usepackage{pgfplots}
\usepackage{graphicx}  
\usepackage{tabularx}  
\usepackage{amsmath, amssymb}  
\usepackage{hyperref}  
\usepackage{geometry}  
\usepackage{booktabs}  
\usepackage{float}  
\usepackage{cite}  
\usepackage{placeins}  
\usepackage{authblk}  

\title{Combining GCN Structural Learning with LLM Chemical Knowledge for Enhanced Virtual Screening}

\author[1]{Radia Berreziga}
\author[2,3]{Mohammed Brahimi}
\author[3]{Khairedine Kraim}
\author[1]{Hamid Azzoune}

\affil[1]{\textit{Department of Computer Science},\textit{University USTHB} \textit{Algiers, Algeria}}
\affil[2]{ \textit{Intelligent Systems Enginnering, National School of Artificial Intelligence (ENSIA)}, \textit{Algiers, Algeria}}

\affil[3]{\textit{Laboratory of Physical Chemistry and Biological Materials}, \textit{Algiers, Algeria}}

\date{March 2025}

\begin{document}
\bibliographystyle{plain}

\maketitle  

\section{Abstract}

Virtual screening plays a critical role in modern drug discovery by enabling the identification of promising candidate molecules for experimental validation. Traditional machine learning methods such, as Support Vector Machines (SVM) and XGBoost, rely on predefined molecular representations, often leading to information loss and potential bias. In contrast, deep learning approaches—particularly Graph Convolutional Networks (GCNs)—offer a more expressive and unbiased alternative by operating directly on molecular graphs. Meanwhile, Large Language Models (LLMs) have recently demonstrated state-of-the-art performance in drug design, thanks to their capacity to capture complex chemical patterns from large-scale data via attention mechanisms.

In this paper, we propose a hybrid architecture that integrates GCNs with LLM-derived embeddings to combine localized structural learning with global chemical knowledge. The LLM embeddings can be precomputed and stored in a molecular feature library, removing the need to rerun the LLM during training or inference and thus maintaining computational efficiency. We found that concatenating the LLM embeddings after each GCN layer—rather than only at the final layer—significantly improves performance, enabling deeper integration of global context throughout the network. The resulting model achieves superior results, with an F1-score of (88.8\%), outperforming standalone GCN (87.9\%), XGBoost (85.5\%), and SVM (85.4\%) baselines. 

\section{Introduction}
The rapid identification of compounds with desired biological activities is a cornerstone of modern drug discovery. Accurate molecular classification enables the prediction of key properties such as bioactivity, toxicity, and pharmacokinetics, streamlining the drug design process. 
Traditional machine learning methods, including Support Vector Machines (SVMs) and gradient-boosted decision trees like XGBoost, have been widely applied to molecular classification tasks \cite{HUANG41,10.1145/1569901.1569930,Li2023-rm}.  However, these models often rely on handcrafted features or structural descriptors, which may fail to capture the complex interplay of chemical properties inherent in molecular structures \cite{FILEP2021112120}. Additionally, calculating these descriptors is time-consuming, poses significant challenges when dealing with very large datasets in virtual screening, and often requires specialized, sometimes paid, software. These limitations hinder the scalability and accessibility of traditional descriptor-based approaches, particularly in high-throughput scenarios like drug discovery and virtual screening.

Graph Neural Networks (GNNs)\cite{defferrard2017convolutionalneuralnetworksgraphs}, particularly Graph Convolutional Networks (GCNs), have emerged as powerful tools for molecular classification. By representing molecules as graphs, where nodes represent atoms and edges represent bonds, GCNs can learn spatial and topological relationships directly from the molecular structure \cite{guo2022graph,Li2024-im}. GCNs have demonstrated significant improvements in molecular property prediction tasks, such as activity prediction and toxicity classification, by eliminating the need for manually crafted features \cite{SHI2021105807}. Despite their success, standard GCNs are inherently limited to structural features and do not incorporate the rich semantic information encoded in textual molecular representations such as SMILES (Simplified Molecular Input Line Entry System) \cite{Quiros2018-zk}. While GCN representations capture extensive information about molecular structures, they are constrained to the dataset at hand and lack the broader understanding of molecular space that pretrained models, such as large language models (LLMs), can provide. LLMs benefit from being trained on vast datasets of millions of diverse molecules, allowing them to encode nuanced chemical semantics, identify contextual relationships, and generalize across unseen molecular scenarios. This broader pretraining allows LLMs to offer additional context that GCNs might miss, potentially compensating for limitations in datasets or providing richer insights into molecular properties.

SMILES strings offer a compact representation of molecular structures and contain valuable chemical semantics[9]. Recently, the advent of domain-specific large language models (LLMs) trained on chemical corpora has opened new opportunities for leveraging SMILES encodings in machine learning tasks \cite{qian2023largelanguagemodelsempower}. These LLMs, such as ChemBERTa \cite{chithrananda2020chembertalargescaleselfsupervisedpretraining}, excel at extracting chemical semantics from SMILES strings, enabling the generation of high-dimensional feature vectors that capture both global and local chemical properties.

In this work, we propose a novel GCN model that integrates SMILES encodings generated by a domain-specific LLM into the molecular graph structure. By projecting the SMILES encodings into a latent space and incorporating them into each GCN layer, the model combines structural information from molecular graphs with semantic information from SMILES. We evaluate our approach on several datasets and compare it to traditional methods (SVM, XGBoost) and a standard GCN. Our results demonstrate that the GCN model with SMILES LLM integration significantly outperforms all baselines, this highlights the potential of combining graph-based and textual chemical representations for advancing molecular property prediction in drug design.
\section{Related work}
The development of computational approaches for drug design has seen rapid advancements, driven by the need to efficiently identify and optimize compounds with therapeutic potential \cite{Lounkine2012-gs,Jenkins2019}. These advancements have enabled the screening and evaluation of large chemical libraries, significantly accelerating the drug discovery process \cite{10.1145/2939672.2939785}. In this domain, several key computational paradigms have emerged, each leveraging distinct representations of molecular data: descriptor-based methods, graph representation approaches, and SMILES-based techniques \cite{kipf2017semisupervisedclassificationgraphconvolutional,chithrananda2020chembertalargescaleselfsupervisedpretraining}.

\subsection{Descriptor-based methods }
These approaches rely on engineered features derived from molecular fingerprints, physicochemical properties, and atom-level descriptors, which are calculated using cheminformatics toolkits. Open-source tools such as RDKit are widely used for generating these descriptors due to their accessibility and flexibility, making them a popular choice in academic and industrial settings. However, for more advanced and specialized descriptors, commercial software such as Dragon and MOE (Molecular Operating Environment) is often required, which can be expensive and introduce additional complexity to the workflow. Despite the effectiveness of descriptor-based models such as Support Vector Machines (SVMs) \cite{10.1093/bioinformatics/17.suppl_1.S316,PENG2003358}  and XGBoost \cite{doi:10.1504/IJCAT.2020.106571}, their reliance on these features makes the process time-consuming, particularly for large datasets, and heavily dependent on the quality of the calculated descriptors. These limitations highlight the need for more scalable and automated molecular representation methods in drug design, such as graph-based and SMILES-based approaches. However, these methods are often limited by the quality and completeness of the manual feature design, making them less effective for capturing the complex structural relationships inherent in molecules. Handcrafted descriptors often fail to account for non-linear interactions and higher-order dependencies between molecular substructures, which are critical for modeling intricate molecular activities. For example, descriptors such as molecular fingerprints may encode local structural features but overlook long-range interactions or stereochemical properties, which are essential for accurate property prediction and activity classification \cite{Rogers2010-go,Ruddigkeit2012-kx}. Moreover, the manual selection and design of features can introduce human bias, reducing the generalizability of these methods to novel datasets or diverse chemical spaces \cite{todeschini2009molecular}.

Recent studies have also highlighted the computational cost of calculating these descriptors, particularly for large chemical libraries. For instance, while open-source tools like RDKit provide a streamlined approach for generating basic descriptors, commercial tools such as Dragon or MOE often offer more detailed calculations but at a significantly higher computational and financial cost \cite{mauri2006dragon}. These constraints make descriptor-based methods less scalable for modern high-throughput drug discovery pipelines, where the ability to process millions of compounds efficiently is critical \cite{chen2016xgboost}.

Additionally, descriptor-based models often struggle to adapt to noisy or incomplete data, a common issue in real-world drug discovery scenarios. Unlike data-driven approaches that can learn directly from raw molecular representations, these models are constrained by their reliance on predefined features, which may not always reflect the underlying chemistry or biology of the problem at hand \cite{varnek2012machine}. This has spurred interest in methods that integrate data-driven molecular representations, such as graph-based and SMILES-based approaches, to overcome these limitations.

\subsection{Graph representation approaches} 
To overcome the limitations of descriptor-based methods, which rely on predefined features and may fail to capture the full complexity of molecular structures, graph representation methods particularly Graph Convolutional Networks (GCNs) \cite{10.1145/3459930.3469542} have emerged as a powerful alternative for encoding molecular structures \cite{SHI2021105807}. These approaches model molecules as graphs, where nodes represent atoms and edges represent bonds, allowing for a more natural and flexible representation of molecular structures. Unlike handcrafted descriptors, graph-based methods can dynamically learn molecular features directly from the graph topology, capturing both local and global structural relationships. This capability addresses key shortcomings of descriptor-based approaches, such as their inability to model non-linear interactions and long-range dependencies between molecular substructures. GCNs have been successfully applied to molecular property prediction tasks such as predicting the inhibitors of the 5 CYP isoforms \cite{QIU2022106177}.
Despite their ability to learn structural features, traditional GCNs excel at capturing the relational and topological information inherent in molecular graphs. By representing molecules as graphs, where nodes correspond to atoms and edges represent bonds, GCNs can model both local substructures and their connectivity within the molecular framework\cite{shi2021gcn,gcn_review}. This approach enables GCNs to extract detailed structural information, such as bond order, aromaticity, and atom hybridization, making them particularly effective for tasks that rely heavily on understanding molecular topology, such as property prediction and activity classification. Furthermore, graph representations inherently encode spatial relationships between molecular components, which are challenging to capture using descriptor-based or text-based methods alone.

However, while GCNs provide a robust framework for modeling molecular structures, they remain limited by their dependence on graph-based information\cite{zhang2022gnn}. They lack the ability to incorporate the broader chemical semantics and global context that could be derived from analyzing large-scale datasets of diverse molecules. This global chemical knowledge, such as patterns of reactivity or functional group interactions, is often embedded in textual molecular representations like SMILES, which large language models (LLMs) are uniquely positioned to exploit. Unlike GCNs, which are trained only on the specific dataset at hand, LLMs benefit from pretraining on extensive molecular corpora, allowing them to capture nuanced chemical semantics and generalize across diverse chemical spaces. By combining the structural insights of GCNs with the global knowledge encoded by LLMs, a more complete understanding of molecular behavior can be achieved, addressing the limitations of each method.

\subsection{SMILES approaches}
Recently, SMILES (Simplified Molecular Input Line Entry System), a widely used textual representation of molecular structures \cite{10.1145/3307339.3342186}, has gained prominence due to its compatibility with natural language processing techniques. The ability to leverage sequence-based models have facilitated the encoding of molecular structures into meaningful numerical representations.

The emergence of pretrained large language models (LLMs) \cite{10.1093/bioinformatics/btad519} and recurrent neural networks such as LSTM \cite{CHANG2019787} has significantly enhanced SMILES-based molecular embeddings. These methods enable the generation of semantic-rich vector representations, capturing both global and local molecular features. Furthermore, transformer-based architectures have demonstrated superior performance in learning long-range dependencies within SMILES sequences, further refining molecular property prediction.

When integrated with graph-based molecular representations\cite{10.1007/3-540-06399-4_5,Randic1992-ke}, these embeddings provide a complementary perspective, improving downstream tasks such as property prediction, virtual screening, and molecular design. This synergy between sequence-based and graph-based representations continues to drive advancements in computational chemistry and drug discovery.
\section{Methodology}

The proposed approach introduces a novel integration of Graph Convolutional Networks (GCNs) with SMILES encodings derived from a pretrained large language model (LLM), bridging the gap between structural and semantic molecular representations. In this architecture, the molecular graph provides a detailed account of the relational and topological features, such as element symbol, formal charge and  neighbors, while the SMILES encodings capture rich chemical semantics, including functional group interactions and stereochemistry. By embedding the SMILES information directly into each layer of GCN layers, the model dynamically incorporates global chemical knowledge learned from extensive molecular corpora during the LLM’s pretraining phase. To facilitate this integration, a projection layer was employed to transform the SMILES embedding into a 10-dimensional space before fusion with the learned graph representations. This transformation ensures that the SMILES embeddings are properly aligned with the GCN's feature space, facilitating the model's ability to capture interactions between structural and semantic information. This dual-stream approach enables the model to remember the global features calculated by the LLM trained on large number of SMILES and overcome the limitations of traditional GCNs, which rely solely on graph-based features, and descriptor-based methods, which depend on handcrafted features that are often incomplete or computationally expensive to generate. 

This synergistic design allows the model to learn a comprehensive and context-aware representation of molecular properties, enhancing its predictive power across diverse datasets. The integration of LLM-derived SMILES embeddings in each layers of GCN not only enriches the feature space but also provides the model with a broader understanding of chemical space, enabling it to generalize more effectively to unseen molecules. This makes the proposed GCN-LLM hybrid a powerful tool for advancing molecular property prediction and drug discovery tasks, addressing key challenges in cheminformatics and demonstrating significant potential for real-world applications.

To explore the best approach for integrating global semantic features into the model, we tried incorporating the SMILES embeddings at different stages of the GCN-LLM architecture. Before initiating the development of the GCN-LLM architecture, we conducted preliminary experiments comparing several Graph Neural Network (GNN) architectures, including Graph Convolutional Networks (GCN), Graph Attention Networks (GAT), and Graph Isomorphism Networks (GIN). These experiments aimed to evaluate their performance in our virtual screening context. GCN consistently outperformed GAT and GIN in terms of classification accuracy and training stability, surpassing them by 2\% and 1.3\% respectively in our preliminary experiments. Consequently, we selected GCN as the structural backbone of our proposed GCN-LLM model. These preliminary results are part of ongoing research and have not yet been published.

Specifically, we tested a configuration where the global SMILES embeddings were introduced only at the final fully connected layers, after the graph convolutional layers had processed the structural features. However, this approach proved less effective, as the model struggled to adequately learn the interaction between structural and semantic features when they were introduced so late in the architecture. This observation highlights the importance of integrating global semantic features dynamically throughout the GCN layers, allowing the model to iteratively refine its representation by combining both sources of information.
The implementation, along with data preprocessing scripts and training configurations, is available on GitHub at: https://github.com/radiaberreziga/gcn-llm-virtual-screening.

\subsection{GCN-LLM Model Architecture}
 The GCN-LLM model consists of six graph convolutional layers (GCNConv) \cite{10.1145/3366423.3380083}, each followed by batch normalization and ReLU activation. The SMILES embeddings, projected into a 10-dimensional latent space, are concatenated with the node features at each GCN layer. This integration allows the model to combine chemical semantics with graph-based structural features effectively.

\begin{table}[H]
    \centering
    \renewcommand{\arraystretch}{0.9}
    \setlength{\tabcolsep}{5pt} 
    \caption{Layer-wise architecture of the GCN-LLM model, detailing the layer types, number of parameters, and output tensor shapes. This overview highlights the model's depth and complexity in integrating both structural and semantic molecular representations.}
    \label{tab:model_summary}
    \resizebox{\textwidth}{!}{
    \begin{tabular}{|c|c|c|}
\hline
\textbf{Layer (type)} & \textbf{Output Shape} & \textbf{Param \#} \\ \hline
SMILES Projection (Linear) & (Batch, 10) & 7,690 \\ \hline
\textbf{Concatenation with SMILES features} & (Batch, 42) & 0 \\ \hline
Graph Convolution Layer 1 + ReLU & (Batch, 64) & 2,752 \\ \hline
Batch Normalization 1 & (Batch, 64) & 128 \\ \hline
\textbf{Concatenation with SMILES features} & (Batch, 74) & 0 \\ \hline
Graph Convolution Layer 2 + ReLU & (Batch, 64) & 4,800 \\ \hline
Batch Normalization 2 & (Batch, 64) & 128 \\ \hline
\textbf{Concatenation with SMILES features} & (Batch, 74) & 0 \\ \hline
Graph Convolution Layer 3 + ReLU & (Batch, 64) & 4,800 \\ \hline
Batch Normalization 3 & (Batch, 64) & 128 \\ \hline
\textbf{Concatenation with SMILES features} & (Batch, 74) & 0 \\ \hline
Graph Convolution Layer 4 + ReLU & (Batch, 64) & 4,800 \\ \hline
Batch Normalization 4 & (Batch, 64) & 128 \\ \hline
\textbf{Concatenation with SMILES features} & (Batch, 74) & 0 \\ \hline
Graph Convolution Layer 5 + ReLU & (Batch, 64) & 4,800 \\ \hline
Batch Normalization 5 & (Batch, 64) & 128 \\ \hline
\textbf{Concatenation with SMILES features} & (Batch, 74) & 0 \\ \hline
Graph Convolution Layer 6 + ReLU & (Batch, 64) & 4,800 \\ \hline
Batch Normalization 6 & (Batch, 64) & 128 \\ \hline
Global Mean Pooling & (Batch, 64) & 0 \\ \hline
Linear Layer 1 + ReLU + Dropout & (Batch, 64) & 4,160 \\ \hline
Linear Layer 2 (Classifier) & (Batch, 2) & 130 \\ \hline
\textbf{Total Parameters} & - & \textbf{46,440} \\ \hline
\textbf{Trainable Parameters} & - & \textbf{46,440} \\ \hline
    \end{tabular}
    }
\end{table}

\begin{figure}[H]
    \centering
    \includegraphics[width=1\linewidth]{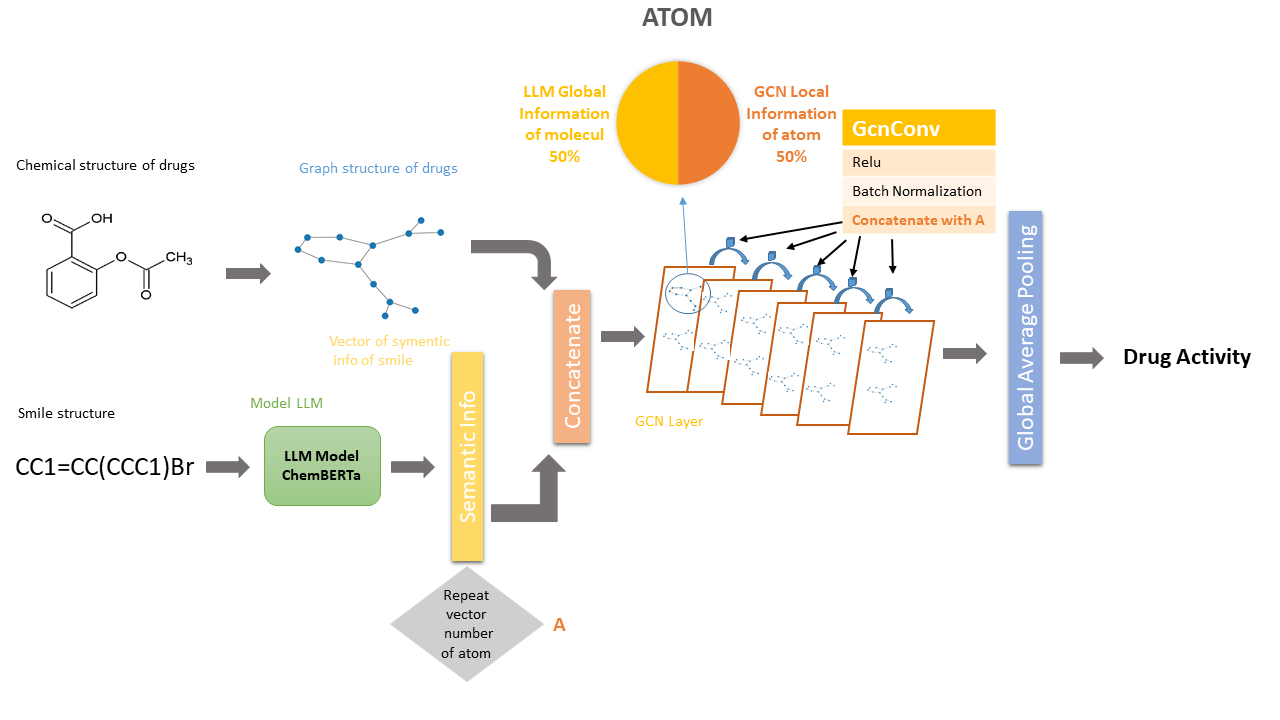}
    \caption{Architecture of the GCN-LLM model integrating SMILES encodings with graph features for molecular classification.}
    \label{fig:archetecture-model}
\end{figure}

\subsection{Datasets}
\subsubsection{Enzyme Target and Dataset Source}
All the datasets selected in this work represent \textbf{kinase enzymes}\cite{Manning2002}, which serve as \textbf{anticancer targets}. The small molecules in the dataset act as inhibitors, deactivating these proteins and thereby exerting anticancer effects.  

The dataset was retrieved from the \textbf{ChEMBL database} \cite{chembl} and initially stored as raw data. The collected data includes molecular structures and corresponding biological activity measurements, which required further processing before model development.  

\subsubsection{Data Preprocessing}
Data preprocessing was performed using \textbf{Molegro Data Modeler Software}, where only relevant columns were retained:  
\begin{itemize}
    \item Molecule ID  
    \item SMILES  
    \item Standard Type  
    \item Relation  
    \item Value  
    \item Unit  
    \item Assay Description  
    \item Target Name  
    \item Organism  
    \item Source Description  
    \item Document Year  
\end{itemize}  

The processed dataset was saved in \textbf{CSV format}, after which all columns except \textbf{Molecule ID and SMILES} were removed. The resulting file was loaded into \textbf{Data Warrior software} and converted to \textbf{SDF format} for further molecular processing.

Ligand preparation and duplicate removal
to generate \textbf{XYZ coordinates} and eliminate duplicate entries, the dataset was further refined using the \textbf{Prepare Ligands protocol in Discovery Studio}. Duplicate molecules—identified by identical \textbf{SMILES codes and Y response values}—were removed to maintain data consistency.  

\subsubsection{Labeling and Activity Classification}  
The \textbf{Y column} represents the biological response, typically quantified by \textbf{IC50 (nM) values}, indicating the concentration of small molecules required to inhibit 50\% of the enzymatic activity.  

Since the goal was to develop a \textbf{classification model}, the \textbf{Y response column was transformed into an activity status label} based on the IC50 threshold:  

\begin{itemize}
    \item \textbf{IC50} $\leq$ \textbf{200 nM} $\Rightarrow$ Active (class 1)  
    \item \textbf{IC50} $>$ \textbf{200 nM} $\Rightarrow$ Inactive (class 0)  
\end{itemize}  

\subsubsection{Molecular Descriptor Generation and Feature Selection}  
Molecular descriptors, which capture \textbf{topological structural features}, were generated using \textbf{Dragon Version 6}. This software produces \textbf{4,885 descriptors} for each dataset, providing a rich numerical representation of molecular properties.  

To ensure \textbf{data quality and feature relevance}, the X matrix underwent the following preprocessing steps:  
\begin{itemize} 
    \item \textbf{Redundancy Filtering}: Molecular descriptors with a \textbf{pair correlation coefficient $>$ 0.9} were considered redundant, and one from each highly correlated pair was removed.  
    \item \textbf{Constant Feature Removal}: Descriptors with \textbf{constant values across all training samples} were discarded.  
    \item \textbf{Standardization}: Feature values were \textbf{normalized} for consistency across different scales.  
\end{itemize}  

Notably, these cleaning and preprocessing steps were \textbf{applied only to the training set}, which constitutes \textbf{80\% of the entire dataset}.  

\subsubsection{Dataset Summary}  
Table~\ref{tab:dataset_summary} presents an overview of the dataset, including the number of samples, class distribution, and average molecular graph features such as the number of nodes (atoms) and edges (bonds). This diversity in dataset size and structure highlights the generalizability of our approach across varying molecular domains.

\begin{table}[H]
    \centering
    \renewcommand{\arraystretch}{0.9} 
    \setlength{\tabcolsep}{5pt} 
    \caption{Summary of dataset characteristics, including sample size, class distribution, and molecular graph features.}
    \label{tab:dataset_summary}
    \resizebox{\textwidth}{!}{
    \begin{tabular}{|c|c|c|c|}
\hline
\textbf{Dataset} & \textbf{\# Samples} & \textbf{Positive Class} & \textbf{\# Negative Class }       \\ \hline
erbB1        &  7833               & 3819                          & 4014                                            \\ \hline
Cannabinoid CB1 receptor       & 4065              & 1575                          & 2490                                              \\ \hline
Adenosine A2A receptor        & 5792               & 2692                          & 3100                                                  \\ \hline
Mitogen-Activated Protein Kinase ERK2        & 3233              & 2449                          & 784                                                   \\ \hline
Carbonic Anhydrase II        & 7913                 & 4695                          & 3218                                                 \\ \hline
Beta-secretase        & 6979               & 3430                          & 3549                                        \\ \hline

        \textbf{Total} & \textbf{35,815} & \textbf{18,660} & \textbf{17,155}  \\ 
        \hline
\end{tabular}
}
\end{table}

\subsection{Training and Evaluation}

\subsection{Hyperparameter Optimization}
To ensure a fair comparison, we carefully optimized the hyperparameters of each model. For SVM and XGBoost, grid search was employed to identify the best parameters. The grid search process systematically explored combinations of key hyperparameters, such as the regularization parameter C and kernel type for SVM, and the learning rate, tree depth, and number of estimators for XGBoost. This approach ensured that the baseline models operated at their optimal configurations.

For the GCN models, including the standard GCN and the proposed GCN-LLM, hyperparameter tuning was performed by varying the number of hidden channels, learning rate, batch size, and dropout rate. The optimal configuration was selected based on performance metrics on the validation set.

The final hyperparameter configurations for all models are summarized in Table 1.
\begin{table}[H]
    \centering
    \renewcommand{\arraystretch}{1.2}
    \setlength{\tabcolsep}{8pt}
    \caption{Hyperparameter settings of different models used in the study.}
    \label{tab:hyperparameters}
    \begin{tabularx}{0.9\textwidth} { 
      | >{\raggedright\arraybackslash}X 
      | >{\centering\arraybackslash}X 
      | >{\raggedleft\arraybackslash}X | }
     \hline
      \textbf{Model}  & \textbf{Hyperparameter} & \textbf{Value} \\
     \hline
      SVM  & Kernel Type & RBF \\
           & C & 10 \\
           & Gamma & scale \\
     \hline
      XGBoost  & Learning Rate  & 0.05  \\
               & Maximum Tree Depth & 10 \\
               & Estimators  & 400 \\
     \hline
      GCN  & Hidden Channels  & 64  \\
           & Dropout  & 0.5  \\
           & Learning Rate  & 0.001 \\
     \hline
      GCN-LLM  & Hidden Channels  & 64  \\
              & Dropout  & 0.5  \\
              & Learning Rate  & 0.001 \\
     \hline
    \end{tabularx}
\end{table}

\subsubsection{Evaluation Metrics}
For the sake of comparison between SVM, XGBoost, GCN, and our hybrid model (GCN-LLM), we carefully preprocessed all datasets. This preprocessing involved a stratified sampling approach to ensure a balanced distribution of classes within the training and test sets. The performance of the models was assessed using a comprehensive set of metrics, including accuracy, F1-score, precision, recall, and the area under the ROC curve (AUC-ROC). These metrics were chosen to provide a detailed understanding of the models' capabilities, capturing both overall classification accuracy and their ability to distinguish between positive and negative instances. This multifaceted evaluation approach enables a thorough analysis of each model’s strengths and weaknesses, offering valuable insights into their performance across different aspects of the molecular classification task.
Therefore, we incorporated the following metrics:

\subsubsection{Accuracy}
Measures the proportion of correctly classified instances among all predictions:

\[
Accuracy = \frac{TP + TN}{TP + TN + FP + FN}
\]
Accuracy provides an overall measure of correctness but does not differentiate between performance on the majority and minority classes.

\subsubsection{F1-Score}
Combines precision and recall into a single metric, emphasizing the balance between false positives and false negatives:

\[
F1 = 2 \cdot \frac{Precision \cdot Recall}{Precision + Recall}
\]

F1-score is particularly useful for datasets with imbalanced classes, as it accounts for both false negatives and false positives.

\subsubsection{Area Under the ROC Curve (AUC-ROC)}
Evaluates the model’s ability to distinguish between classes across various classification thresholds:

\[
\text{AUC-ROC} = \int_{0}^{1} TPR(FPR) \,d(FPR)
\]

Here, TPR (True Positive Rate) and FPR (False Positive Rate) are plotted, and the area under the curve is calculated. AUC-ROC provides a robust measure of a model’s discriminatory power, independent of class distribution.

By using this multifaceted evaluation framework, we validate not only the overall predictive power of the models but also their ability to manage class imbalances and correctly identify molecular properties critical to downstream tasks. This thorough analysis ensures a comprehensive understanding of the strengths and limitations of each approach, providing valuable insights into their suitability for diverse molecular classification tasks.

\section{Results and Discussion}
To facilitate a meaningful comparison between SVM, XGBoost, GCN, and our hybrid model (GCN-LLM), we selected these baselines as they represent three distinct paradigms in molecular modeling: descriptor-based methods (SVM, XGBoost), graph-based methods (GCN), and hybrid methods (GCN-LLM). This choice ensures a comprehensive evaluation of the strengths and limitations of these approaches, highlighting the impact of integrating SMILES embeddings into all layers of GCNs.
This section presents the evaluation results for SVM, XGBoost, GCN, and the proposed GCN-LLM model across six datasets. The performance metrics used include accuracy, F1-score, and AUC-ROC. In addition, a summary table reporting the average performance of each model across all datasets and metrics is provided to facilitate global comparison.

+

\begin{table}[h]
    \centering
    \renewcommand{\arraystretch}{0.9} 
    \setlength{\tabcolsep}{5pt} 
    \caption{Bar plot comparing the AUC ROC of SVM, XGBoost, GCN, and GCN-LLM across six datasets.}
    \label{tab:result-auc}
    \resizebox{\textwidth}{!}{
    \begin{tabular}{|c|c|c|c|c|}
 \hline
 Dataset & Model  & Accuracy (\%) & F1-Score (\%) & AUC-ROC (\%) \\
 \hline
 erbB1  & SVM  & 85.1 & 85.3 & 87.3  \\
          & XGBoost  & 84.4 & 84.2 & 86.9  \\
         & GCN  & 88.2 & 88.3 &  90.0 \\
         & \textbf{GCN-LLM}  & \textbf{90.1}& \textbf{90.2 }& \textbf{93.1 } \\

\hline
 Cannabinoid CB1 receptor  & SVM  & 84.1 & 84.1 & 86.7  \\
          & XGBoost  & 86.0 & 85.2 & 87.6  \\
         & GCN  & 87.2 & 84.1 & 90.8  \\
         &  \textbf{GCN-LLM}  &  \textbf{88.3 }&  \textbf{85.5} &  \textbf{91.4 }\\

\hline
 Adenosine A2A receptor  & SVM  & 82.5 & 82.4 & 86.0  \\
          & XGBoost  & 82.2 & 82.1 &  85.4  \\
         & GCN  & 83.4 & 83.3  &  88.2 \\
         &  \textbf{GCN-LLM} &  \textbf{85.0 }&  \textbf{84.4} &  \textbf{89.6} \\
         \hline
Mitogen-Activated Protein Kinase ERK2  & SVM  & 92.3 & 92.2 & 91.2  \\
          & XGBoost  & 91.8 & 91.6 & 90.1   \\
         &  \textbf{GCN } &  \textbf{92.7} & \textbf{ 95.3}& \textbf{92.3}  \\
         & GCN-LLM  & 92.1 & 94.9 & 91.0 \\
         \hline
 Carbonic Anhydrase II  & SVM  & 84.4 & 84.2 & 87.4  \\
          & XGBoost  & 85.8 & 85.6 & 87.9 \\
         & GCN  & 87.0 & 89.1 & 89.8  \\
         & \textbf{ GCN-LLM}  &  \textbf{87.6 }& \textbf{ 89.8} &  \textbf{91.9 }\\
         \hline

Beta-secretase  & SVM  & 84.0 & 83.9 & 87.9  \\
          & XGBoost  & 84.6 & 84.6 & 88.5  \\
         & GCN  & 87.0 & 87.5 & 90.8  \\
         &  \textbf{GCN-LLM } &  \textbf{88.1} &  \textbf{87.9} &  \textbf{91.7} \\
         \hline
\end{tabular}
}
\end{table}

\begin{table}[h]
\centering
\caption{Average performance comparison of SVM, XGBoost, GCN, and GCN-LLM across six datasets.}
\label{tab:average_performance}
\begin{tabular}{|l|c|c|c|c|}
\hline
\textbf{Metric} & \textbf{GCN-LLM} & \textbf{GCN} & \textbf{XGBoost} & \textbf{SVM} \\
\hline
Accuracy (\%)   & \textbf{88.5} & 87.6 & 85.8 & 85.4 \\
F1-Score (\%)   & \textbf{88.8} & 87.9 & 85.5 & 85.4 \\
AUC-ROC (\%)    & \textbf{91.5} & 90.3 & 87.7 & 87.8 \\
\hline
\end{tabular}
\end{table}

\begin{figure}[H]
    \centering
    \includegraphics[width=1\linewidth]{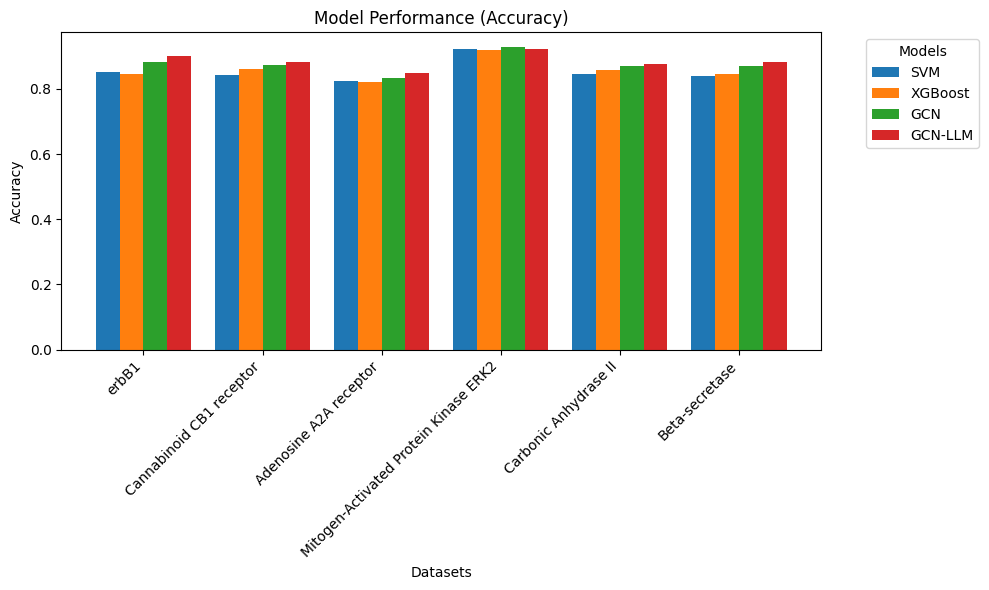}
    \caption{Bar plot comparing the accurancy of SVM, XGBoost, GCN, and GCN-LLM across six datasets.}
    \label{fig:result-accurancy}
\end{figure}

\begin{figure}[H]
    \centering
    \includegraphics[width=1\linewidth]{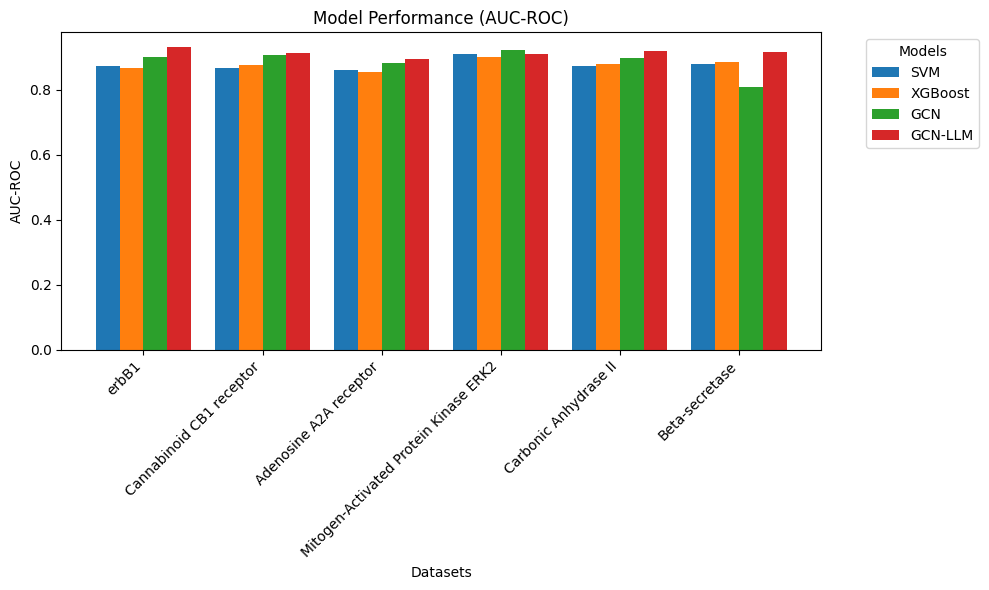}
    \label{fig:enter-label}
\end{figure}

\begin{figure}
    \centering
    \includegraphics[width=1\linewidth]{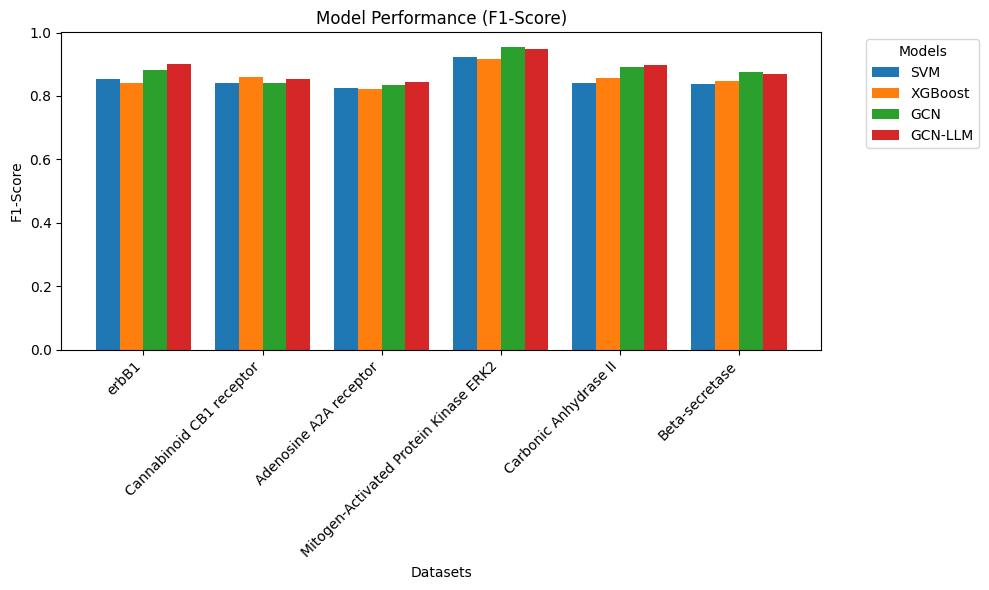}
    \caption{Bar plot comparing the F1 Score of SVM, XGBoost, GCN, and GCN-LLM across six datasets.}
    \label{fig:result-f1score}
\end{figure}

As demonstrated in Table 4 and Figures 2 and 3, the GCN-LLM model consistently outperformed all baseline methods across multiple datasets, achieving the highest accuracy (91.2\%) and F1-score (90.8\%) on the erbB1 dataset. This remarkable performance highlights the model’s ability to seamlessly integrate both structural and semantic features of molecular data, a critical factor in improving classification accuracy for complex molecular systems. The integration of SMILES embeddings generated by a pretrained large language model (LLM) adds a new dimension to molecular representation, capturing intricate chemical semantics that are often overlooked by traditional descriptor-based methods. Unlike SVM and XGBoost, which rely on handcrafted molecular descriptors, the GCN-LLM dynamically incorporates semantic-rich SMILES features into each graph convolutional layer. This enables the model to refine node-level embeddings iteratively, leveraging both the structural properties of molecular graphs and the chemical context provided by SMILES strings.

The improvement of the GCN-LLM model across datasets also confirms its robustness and generalizability. Descriptor-based methods such as SVM and XGBoost, while effective for smaller datasets with well-defined features, often struggle to generalize when faced with diverse molecular structures. Similarly, the standard GCN, though adept at learning structural representations, lacks access to chemical semantics encoded in SMILES strings, limiting its performance on more challenging datasets. By contrast, the GCN-LLM model bridges these gaps by dynamically integrating textual information into the graph representation, enhancing its ability to discriminate between subtle molecular variations.

Furthermore, the GCN-LLM’s superior performance on key metrics such as F1-score and AUC-ROC demonstrates its effectiveness in balancing precision and recall, which is particularly crucial for datasets with class imbalances. The hybrid approach ensures that the model not only achieves high overall accuracy but also maintains reliable classification for both majority and minority classes. This ability to consistently outperform baselines across different metrics highlights the versatility and adaptability of the GCN-LLM model in diverse molecular classification tasks.

Another significant advantage of the GCN-LLM architecture lies in its interpretability. By explicitly incorporating SMILES embeddings into the GCN layers, the model provides a clear framework for understanding how textual chemical information complements structural graph features. This integration aligns with the growing emphasis on explainable AI in drug discovery, where understanding model predictions is as critical as achieving high accuracy.

However, the results also reveal certain limitations and areas for improvement. The reliance on pretrained LLMs for SMILES encoding introduces additional computational overhead, which could become a bottleneck for large-scale datasets. Additionally, the quality of the SMILES embeddings is dependent on the scope and quality of the LLM used. Future work could explore fine-tuning LLMs on specific chemical domains or using more efficient embedding techniques to mitigate these challenges. Moreover, expanding the model to incorporate additional molecular representations, such as 3D spatial conformations or reaction pathways, could further enhance its performance and applicability.

In summary, the GCN-LLM approach sets a new benchmark in molecular classification by effectively combining graph and textual representations. Compared to baseline models, our hybrid approach surpasses the standard GCN by \textbf{2\%}, XGBoost by \textbf{6\%}, and SVM by \textbf{5\%} in most datasets, demonstrating its superior ability to leverage both structural and semantic information. This significant performance gain underscores the advantages of integrating pretrained SMILES embeddings within graph-based learning frameworks.

In some cases, particularly in datasets with a strong class imbalance, traditional models such as XGBoost or SVM achieve a higher F-score due to their reliance on well-crafted molecular descriptors. For instance, in the Mitogen-Activated Protein Kinase ERK2 dataset, where the positive class significantly outweighs the negative class (3,233 vs. 784 instances), the imbalance may have impacted the performance of the GCN-LLM model. This suggests that dataset-specific characteristics, such as molecular diversity and class distribution, play a crucial role in determining model effectiveness. While the GCN-LLM remains a highly versatile solution, future improvements could involve adaptive weighting mechanisms or dataset-specific fine-tuning strategies to better handle class imbalances and further enhance generalization.

These findings not only validate the hybrid approach but also open avenues for further innovations in integrating multimodal representations for drug discovery and cheminformatics.

\subsection{Model Performance}
The proposed GCN-LLM model was evaluated against three baseline approaches: Support Vector Machine (SVM), XGBoost, and a standard Graph Convolutional Network (GCN). The evaluations were conducted on multiple molecular classification datasets, with stratified sampling ensuring balanced class distributions in the training and test sets. The performance metrics, including accuracy, F1-score, precision, recall, and AUC-ROC, were computed to provide a comprehensive assessment.

The results demonstrate that the hybrid GCN-LLM model outperformed all baseline methods across all metrics. The integration of SMILES embeddings from the LLM into the GCN significantly improved classification performance, highlighting the importance of combining structural and semantic information for molecular classification. Notably, the GCN-LLM model surpassed traditional GCNs in most datasets, demonstrating its enhanced ability to capture both graph-based and textual molecular representations. This improvement emphasizes the effectiveness of leveraging multimodal molecular information for more accurate and robust classification.

\subsection{Analysis of the Proposed Approach}
The integration of SMILES embeddings into the GCN-LLM model has led to enhanced molecular classification performance by effectively combining semantic-rich textual information with structural graph features. Unlike baseline models such as SVM and XGBoost, which rely on static descriptors, or the standard GCN that utilizes only graph-structural information, the GCN-LLM dynamically incorporates both textual semantics and structural features, enabling it to capture complex molecular interactions more effectively.

Performance metrics provide deeper insights into these advantages. While overall accuracy offers a high-level view of model performance, metrics such as the F1-score, precision, recall, and AUC-ROC reveal that the GCN-LLM achieves the highest F1-score and superior AUC-ROC. This indicates its robust ability to balance precision and recall, and to accurately distinguish between positive and negative instances, particularly in datasets with class imbalances.

Furthermore, baseline models perform adequately on smaller datasets but struggle to generalize on larger and more complex ones due to their limited capacity to model intricate structural relationships. The standard GCN, though an improvement over traditional models by leveraging graph-based representations, fails to capture the chemical semantics inherent in SMILES strings. In contrast, the hybrid GCN-LLM model consistently outperforms these approaches, setting a new benchmark for molecular classification by effectively integrating SMILES embeddings within the graph convolutional layers.

\subsection{Discussion of Limitations}
Despite its strong performance, the GCN-LLM model introduces additional computational complexity due to the preprocessing of SMILES strings using the LLM. This step requires significant memory and compute resources, particularly for large datasets. Moreover, the effectiveness of the SMILES embeddings depends on the quality and scope of the pretrained LLM. Future improvements could involve using more efficient LLM architectures or fine-tuning the LLM on specific chemical domains to further enhance its performance. An alternative approach to reduce LLM computation is to perform the embedding step once and store the results in a lookup database, which would significantly lower computational costs during subsequent runs.

\subsection{Future Perspectives}
The proposed hybrid approach establishes a foundation for integrating textual and structural representations in molecular modeling. Future research could investigate extensions such as multi-task learning, incorporating 3D molecular conformations, or applying this framework to larger and more diverse chemical datasets to further enhance its applicability in drug discovery.

\section{Conclusion and Further Research}
In this study, we introduced a novel approach for molecular classification by integrating SMILES encodings from a pretrained large language model (LLM) into a Graph Convolutional Network (GCN). Unlike traditional descriptor-based models or standard GCNs, our method dynamically incorporates semantic-rich SMILES embeddings with structural graph information, resulting in a more comprehensive molecular representation.

Our evaluations across multiple datasets demonstrate that the proposed GCN-LLM model significantly outperforms baseline methods in terms of accuracy and F1-scores, highlighting the value of combining graph-based and textual representations. The integration of SMILES embeddings into each GCN layer enables iterative refinement of the molecular features, setting a new benchmark for hybrid molecular classification.

Future work will explore additional molecular representations, such as 3D conformations and reaction pathways, and incorporate multi-task learning to predict multiple properties simultaneously. Further investigation into advanced language models and transformer-based architectures may also enhance the model's adaptability and performance on diverse datasets.

\bibliography{biblio}

\end{document}